\documentclass[sigconf, nonacm]{acmart}

\usepackage{caption}
\usepackage{subcaption}
\usepackage{float}
\usepackage{multirow}
\usepackage{pifont}
\usepackage{makecell}
\usepackage{algorithm}
\usepackage[noend]{algpseudocode}

\usepackage{float}
\usepackage[section]{placeins}

\algrenewcommand\algorithmicrequire{\textbf{Input:}}
\algrenewcommand\algorithmicensure{\textbf{Output:}}

\newcommand{\AX}{\textsc{arxiv}}
\newcommand{\RE}{\textsc{reddit}}
\newcommand{\PA}{\textsc{papers100M}}
\newcommand{\PR}{\textsc{products}}

\newcommand{\kmeans}{\textsc{K-Means}}
\newcommand{\our}{\textsc{EmbedPart}}

\newcommand{\metis}{\textsc{Metis}}
\newcommand{\ldg}{\textsc{Ldg}}
\newcommand{\cuttana}{\textsc{Cuttana}}

\newcommand{\spinner}{\textsc{Spinner}}

\newcommand{\dgl}{\textsc{DGL}}

\newcommand{\otwov}{\textsc{EmbedPart-v-2}}
\newcommand{\otwol}{\textsc{EmbedPart-l-2}}
\newcommand{\othreev}{\textsc{EmbedPart-v-3}}
\newcommand{\othreel}{\textsc{EmbedPart-l-3}}

\usepackage{xcolor} 
\usepackage{graphicx}  
\usepackage{tikz}
\usepackage{xstring} 

\newcommand{\circnum}[2][DAE8FC]{%
  \begingroup
  \StrDel{#1}{\#}[\circnumhex]%
  \definecolor{circnumcol}{HTML}{\circnumhex}%
  \tikz[baseline=(char.base)]{
    \node[
      shape=circle,
      draw=circnumcol,
      fill=circnumcol,
      text=black,
      inner sep=1pt
    ] (char) {#2};
  }%
  \endgroup
}

\begin{document}
\title{\our{}: Embedding-Driven Graph Partitioning for Scalable Graph Neural Network Training}
\author{Nikolai Merkel}
\authornote{Work partially conducted at Technical University of Munich and TU Berlin / BIFOLD.}
\email{nikolai.merkel@tum.de}
\affiliation{%
  \institution{Technical University of Munich (TUM)}
  \city{Munich}
  \country{Germany}
}

\author{Ruben Mayer}
\email{ruben.mayer@uni-bayreuth.de}
\affiliation{%
  \institution{University of Bayreuth}
  \city{Bayreuth}
  \country{Germany}
}

\author{Volker Markl}
\email{volker.markl@tu-berlin.de}
\affiliation{%
\institution{TU Berlin, BIFOLD, DFKI}
\city{Berlin}
\country{Germany}
}
\author{Hans-Arno Jacobsen}
\email{jacobsen@eecg.toronto.edu}
\affiliation{%
  \institution{University of Toronto}
  \city{Toronto}
  \country{Canada}
}

\begin{abstract}
Graph Neural Networks (GNNs) are widely used for learning on graph-structured data, but scaling GNN training to massive graphs remains challenging.
To enable scalable distributed training, graphs are divided into smaller partitions that are distributed across multiple machines such that inter-machine communication is minimized and computational load is balanced.
In practice, existing partitioning approaches face a fundamental trade-off between partitioning overhead and partitioning quality.

We propose \our{}, an \textit{embedding-driven} partitioning approach that achieves both speed and quality.
Instead of operating directly on irregular graph structures, \our{} leverages node embeddings produced during the actual GNN training workload and clusters these dense embeddings to derive a partitioning.

\our{} achieves more than 100$\times$ speedup over \metis{} while maintaining competitive partitioning quality and accelerating distributed GNN training.
Moreover, \our{} naturally supports graph updates and fast repartitioning, and can be applied to graph reordering to improve data locality and accelerate single-machine GNN training.
By shifting partitioning from irregular graph structures to dense embeddings, \our{} enables scalable and high-quality graph data optimization.
\end{abstract}

\maketitle

\vspace{1cm}

\section{Introduction}
Graph Neural Networks (GNNs) have emerged as a state-of-the-art method for extracting insights and predictions from graph-structured data. 
Their application spans diverse domains, including learning on relational databases~\cite{10.1145/3711896.3736558}, recommendation systems~\cite{10.1145/3219819.3219890}, drug discovery~\cite{10.1145/3447548.3467311}, fraud detection~\cite{10.1145/3511808.3557136}, and knowledge graphs~\cite{DBLP:conf/esws/SchlichtkrullKB18}.
However, training GNNs efficiently at scale remains computationally and memory intensive because for each vertex large feature vectors and intermediate representations need to be stored and computationally expensive neural network operations are performed.
Consequently, distributed GNN training has become an essential strategy to scale to large graphs.

\begin{figure}[!t]
\centering
\begin{subfigure}[b]{0.99\linewidth}
\centering
\includegraphics[width=\linewidth]{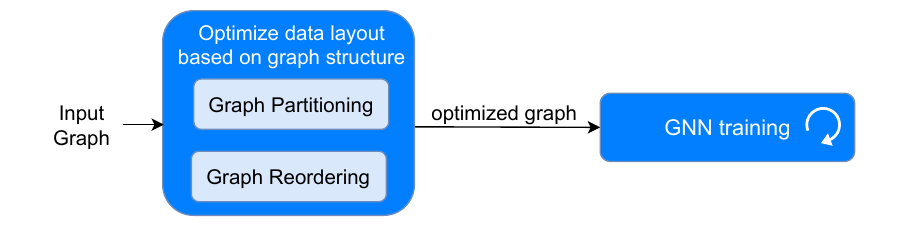}
\caption{Traditional approach.}
\label{fig:overview:traditional}
\end{subfigure}
\begin{subfigure}[b]{0.99\linewidth}
\centering
\includegraphics[width=\linewidth]{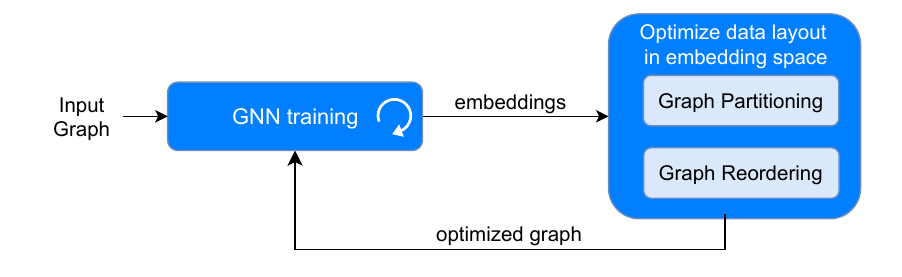}
\caption{Proposed approach: \our{}}
\label{fig:overview:our}
\end{subfigure}
\centering
\caption{(a) Traditional approaches perform graph partitioning or reordering as a preprocessing step based solely on the graph structure.
(b) Our approach (\our{}) derives partitions from node embeddings produced during GNN training.
These embeddings capture both structural and feature information and enable partitioning or reordering directly in the dense embedding space using scalable clustering.}
\label{fig:overview}
\end{figure} 

In order to enable distributed GNN training, the input graph must be partitioned into smaller subgraphs that are distributed across multiple machines. 
Each worker processes the vertices of its local partition; however, for processing, information from vertices on remote partitions may be required. 
Consequently, data such as feature vectors and intermediate representations must be exchanged between machines during distributed training, which introduces substantial communication and synchronization overhead and can dominate end-to-end GNN training time \cite{gnnpartitioing}. 
To reduce the communication burden in distributed graph processing, graph partitioning methods have been proposed~\cite{metis,ldg,spinner,cuttana}, aiming to minimize cross-partition edges that lead to network communication while balancing the number of vertices across machines to achieve workload balance.

\textbf{Challenges.} 
Despite extensive research, existing graph partitioning methods face key limitations:  
(1) they often require expensive upfront computation and partitioning time may not be amortized over the GNN training process,
(2) repartitioning overhead is prohibitive in scenarios where the number of partitions must change dynamically (e.g., when scaling in or out),  
(3) they struggle with efficiently handling updates in dynamic graphs, and
(4) operate on irregular graph data structures which makes it challenging to parallelize, e.g., on a GPU.

\textbf{Solution.}
We propose a novel \emph{embedding-driven graph partitioning} approach, \our{}, that fundamentally differs from classical methods: instead of constructing a partitioning \emph{before} training begins (see Figure~\ref{fig:overview:traditional}), the GNN training is directly started. 
Then, the node embeddings that naturally emerge \emph{during} the GNN workload itself are used to perform graph partitioning in the dense embedding space of the nodes (see Figure~\ref{fig:overview:our}).
\our{} consists of two lightweight phases:
(1) extracting node embeddings produced as a by-product of the GNN training, and
(2) clustering these embeddings in the dense embedding space to derive graph partitions, followed by a lightweight rebalancing step to satisfy \mbox{balancing constraints.}

This design decouples graph partitioning from expensive operations on highly irregular graph structures. Partitioning in the embedding space is efficient and trivially parallelizable: for instance, \kmeans{} can cluster embeddings efficiently in parallel on a GPU.
As a result, partitioning overhead becomes negligible relative to GNN training time and enables fast repartitioning.
Further, \our{} naturally extends to dynamic settings where the graph changes over time, due to the inductive capabilities of GNNs, which can compute embeddings for new vertices.
Additionally, embeddings produced by one specific GNN architecture can be reused to partition graphs for other architectures. 
Finally, thanks to its modular separation of embedding computation, clustering, and balancing, \our{} can be directly applied to graph reordering by reusing the clustering phase while omitting balancing.
This reorders vertices such that nodes within the same partition are placed consecutively in memory, improving cache locality for single-machine GNN training.

\textbf{Contributions.}
Our contributions are summarized as follows:

\begin{itemize}

\item We introduce \our{}, a novel embedding-driven partitioning approach that derives graph partitions from node embeddings produced during the actual GNN training workload.
By shifting partitioning from irregular graph structures to dense embedding spaces, \our{} transforms graph partitioning into a scalable clustering problem and enables efficient parallelization.

\item \our{} has a modular architecture consisting of embedding extraction, clustering, and lightweight balancing.
All components can be replaced; for example, more advanced clustering algorithms could further improve performance.

\item \our{} operates on embeddings that can be reused and computed incrementally, enabling efficient repartitioning for evolving graphs and changing system configurations.
This makes \our{} well suited for dynamic graphs and iterative machine learning workflows where data and resource requirements change over time.

\item We demonstrate that \our{} reduces partitioning overhead by more than two orders of magnitude compared to state-of-the-art in-memory partitioners while maintaining competitive partitioning quality and strong distributed GNN training performance.

\item While primarily designed for graph partitioning, \our{} can also be applied to graph reordering, improving training performance in single-machine settings by increasing cache locality.

\end{itemize}

The paper is organized as follows. 
In Section~\ref{sec:background}, we introduce the graph partitioning problem and provide background on graph neural network (GNN) training. 
Section~\ref{sec:approach} presents \our{}, our novel architecture for graph partitioning. 
In Section~\ref{sec:evaluation}, we evaluate \our{} with respect to partitioning time, partitioning quality, and distributed training performance. We further demonstrate its applicability to graph reordering. 
Section~\ref{sec:relatedwork} discusses related work, and Section~\ref{sec:conclusions} concludes the paper. 

\section{Background}
\label{sec:background}
\subsection{Graph Partitioning}
\begin{figure}[!ht]
\centering
\includegraphics[width=\linewidth]{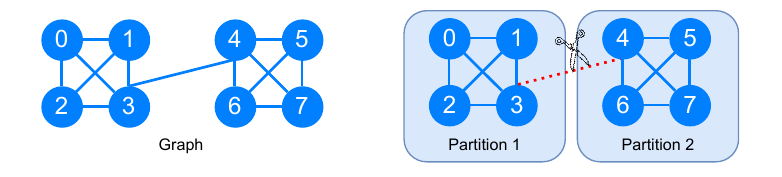}
\caption{Vertices are assigned to partitions. Edges connecting vertices of different partitions are cut.}
\label{fig:partitioning:example:vertexpartitioning}
\end{figure}

Let $G = (V, E)$ be a graph with a set of vertices $V$ and edges $E \subseteq V \times V$. 
Let $N(v)$ be the set of vertices to which vertex $v$ is connected by an edge, also called neighbors of $v$.
In vertex partitioning (see Figure~\ref{fig:partitioning:example:vertexpartitioning}), the set of vertices $V$ is divided in $k$ disjoint partitions $P_1, \dots, P_k$.
By partitioning the vertices, edges may be cut.
For example, in Figure~\ref{fig:partitioning:example:vertexpartitioning}, the edge between vertices 3 and 4 (colored red) is cut.
Each vertex $v \in V$ will be assigned to exactly one partition.
Therefore, $V = \bigcup_{i \in [k]} P_i$ and $P_i \cap P_j = \emptyset$ for $i \neq j$.
Let $\textit{pid(v)}$ be the partition number $p \in [k]$ to which $v$ is assigned. 
An edge $e =(u,v)$ is cut if $\textit{pid(u)} \ne \textit{pid(v)}$ meaning $u$ and $v$ are assigned to different partitions.
\begin{figure*}[!ht]
\centering
\includegraphics[width=\linewidth]{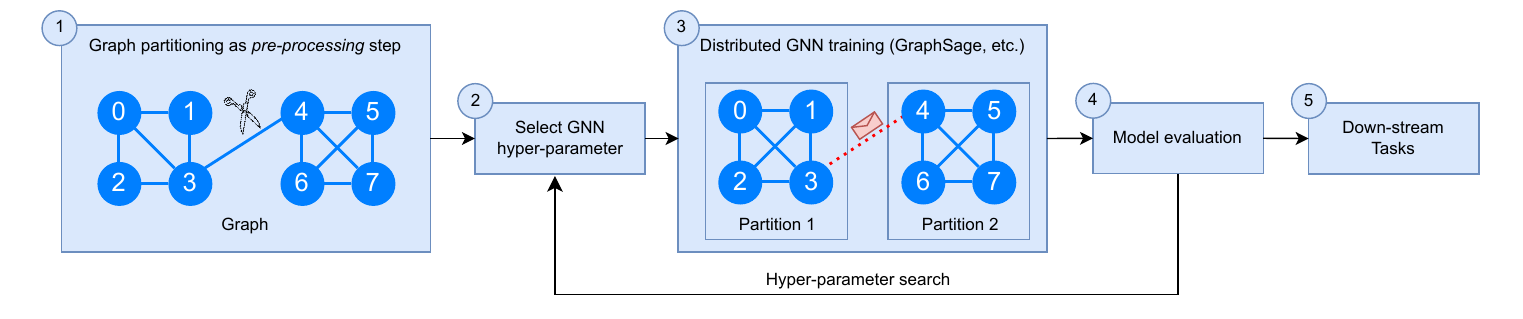}
\caption{Distributed GNN training pipeline: \circnum{1} the input graph is partitioned into $k$ partitions; \circnum{2} hyperparameters for the GNN are selected; \circnum{3} the model is trained in a distributed fashion; \circnum{4} the model is evaluated; repeat \circnum{3}-\circnum{4} with the next hyperparameter set; \circnum{5} once the final model is trained, it is applied to downstream tasks.}
\label{fig:distributed-gnn-pipeline}
\end{figure*}

\subsubsection{Partitioning quality metrics.} 
Two commonly used partitioning quality metrics are the \textit{edge-cut-ratio} and the \textit{vertex balance}.
Let $C = \{(u,v) \in E \; | \;  \textit{pid(u)} \ne \textit{pid(v)}\}$ be the set of edges which are cut.  
The edge-cut-ratio is defined as:
\begin{equation}
\mathit{ECR}=\frac{|C|}{|E|}. 
\end{equation}

The smaller the edge-cut ratio, the fewer edges are cut. 
An edge-cut ratio of 0 indicates that no edges are cut (good), and an edge-cut ratio of 1 indicates that all edges are cut (bad).
The vertex balance of a partitioning $P$ is defined as:
\begin{equation}
\label{eq:vertex_balance}
B_{v}(P)=\frac{\mathit{max}({|P_1|, \dots, |P_k|})}{\mathit{mean}({|P_1|, \dots, |P_k|})}. 
\end{equation}

The vertex balance is between 1 and $k$.
The closer the balance to 1, the better.
A balance of 1 indicates that all partitions have the same number of vertices (good), and a balance of $k$ indicates that one partition contains all vertices while the other partitions are empty (bad).

\subsubsection{Partitioning goal.}
The goal of graph partitioning for distributed GNN training is two-fold:
(i) minimizing the number of cut edges (edge-cut-ratio) and
(ii) balancing the number of vertices across machines (vertex balance).
In distributed GNN training, these partitioning objectives directly impact system performance.
Cut edges induce inter-machine communication during message passing, as feature vectors and intermediate representations must be exchanged across partitions.
At the same time, the computational cost and GPU memory footprint of GNN training scale with the number of local vertices, since features and intermediate representations must be stored and processed per vertex.
Therefore, well-balanced partitions are essential to avoid stragglers and memory bottlenecks, while minimizing cut edges reduces communication overhead.

\subsubsection{Partitioner types.} 
Graph partitioners can broadly be categorized into two classes: 
\textit{in-memory} graph partitioners and \textit{streaming} graph partitioners.
In-memory graph partitioning algorithms load the entire graph into memory before performing the partitioning.
This approach allows to operate with a global view of the graph structure, which typically leads to high-quality partitions (less cut edges) but comes at the cost of significant memory requirements and limited scalability for very large graphs.
In contrast, streaming graph partitioners process the graph in a sequential manner, assigning vertices or edges to partitions as they arrive in the stream.
Since they do not have access to the full graph at once, these methods rely only on local or partial information when making assignment decisions.
While this constraint can lead to lower-quality partitions (more cut edges) compared to in-memory methods, streaming partitioners are considerably more scalable and can handle graphs that are too large to fit into memory, making them attractive for large-scale graph processing scenarios.

The traditional distinction between in-memory and streaming partitioners lies in whether they have access to the full graph structure.
\our{} is largely orthogonal to this distinction.
Since partitioning is performed on dense node embeddings rather than directly on the graph topology, \our{} does not require a global or incremental view of the graph structure during the partitioning step.
Instead, structural information is implicitly encoded in the node embeddings through the message-passing operations performed during GNN training.

\subsection{Graph Neural Network Training}
Graph Neural Networks (GNNs) are a special type of neural network capable of performing deep learning on graph-structured data.
They are widely used for different prediction tasks, such as node classification or link prediction. 

Let $G = (V, E)$ be a graph and $x_v$ be the feature vector of vertex $v$.
Let $h_v^{(l)}$ be the representation of $v$ in layer $l$ and $h_v^{(0)}$ be the feature vector $x_v$.
A GNN consists of $L$ layers, which are iteratively computed. 
In each GNN layer $l$, for a vertex $v$, the hidden representations of its neighboring vertices of the previous layer $l-1$ are aggregated to $a_v^{(l)}$ by applying an aggregation function (see Equation~\ref{eq:1}).
Then, the representation of $v$ is updated based on $a_v^{(l)}$ and $h_v^{(l-1)}$ (see Equation~\ref{eq:2}).

\begin{equation} 
\label{eq:1}
a_v^{(l)} = \mathit{AGGREGATE}^{(l)}(\{h_u^{(l-1)} \mid u \in N(v)\})
\end{equation}
 
\begin{equation} 
\label{eq:2}
h_v^{(l)} = \mathit{UPDATE}^{(l)}(a_v^{(l)}, h_v^{(l-1)})
\end{equation}

The main approaches to train GNNs are \textit{mini-batch} and \textit{full-graph} training~\cite{10.14778/3717755.3717776}.
In mini-batch training, multiple mini-batches are sampled from the graph in each epoch, and each mini-batch leads to a model update. 
In contrast, in full-graph training, in each epoch, the model is updated only once based on the whole graph.

A standard distributed GNN training pipeline is illustrated in Figure~\ref{fig:distributed-gnn-pipeline}. 
It begins with partitioning the input graph across machines~(\circnum{1}), followed by selecting a GNN architecture and corresponding hyperparameters, such as the number of layers or hidden dimensions~(\circnum{2}). 
The selected model is then trained in a distributed manner~(\circnum{3}) and evaluated by measuring the prediction performance with accuracy metrics~(\circnum{4}). 
Based on the prediction performance, different hyperparameters or GNN variants may be explored in iterative retraining cycles. 
Once a satisfactory model is obtained, it is deployed for downstream tasks such as recommendation~(\circnum{5}).

\section{\our{}: Embedding-driven Partitioning}
\label{sec:approach}

\begin{figure*}[!ht]
\centering
\includegraphics[width=\linewidth]{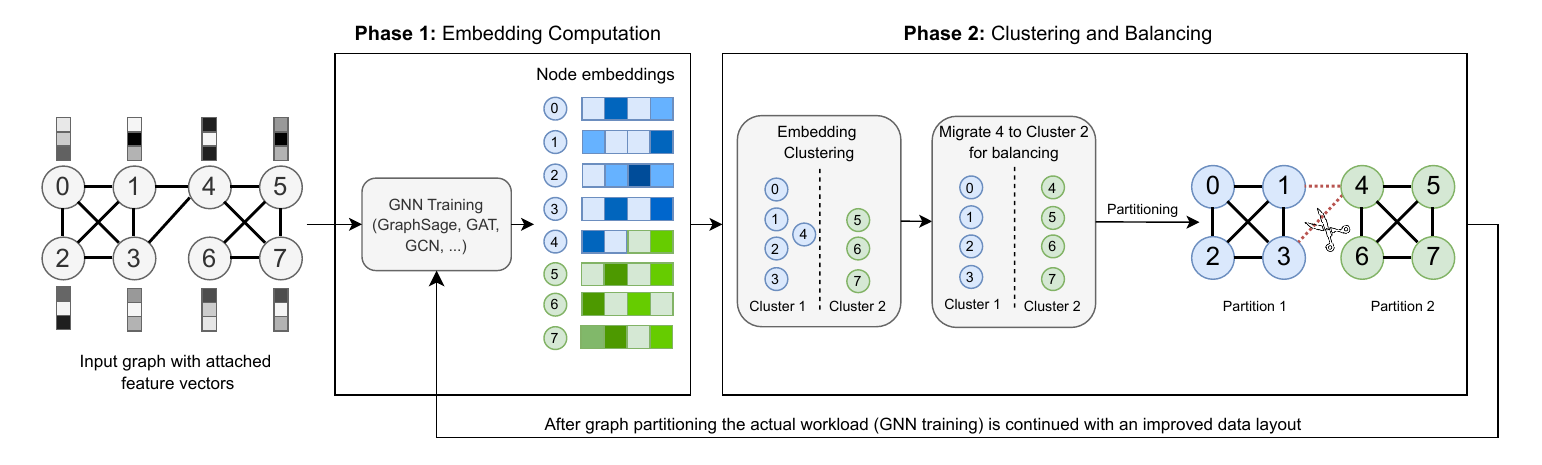}
\caption{\our{} overview: The input to \our{} is like in GNN training, a graph with features that are attached to nodes. In Phase 1, we train a GNN model (the actual workload) and get node embeddings. At any point, the process can transition to Phase 2, where clustering on the embeddings assigns nodes to clusters. To maintain balanced distributed training, nodes are migrated from overloaded to underloaded clusters such as node 4. The resulting clusters are then used for graph partitioning. Finally, the process returns to Phase 1 to continue GNN training with the improved data layout. }
\label{fig:approach}
\end{figure*}

We propose an architecture for \underline{Embed}ding-driven \underline{Part}itioning (\our{}).
The key idea is to leverage node embeddings produced during the actual GNN training workload for graph partitioning. 
Rather than computing a partitioning as a separate pre-processing step, \our{} derives partitions from node embeddings produced during the actual GNN training workload.
These embeddings are then clustered into $k$ partitions, followed by a lightweight balancing step to ensure that partitions satisfy vertex balancing constraints. 
Later on we show how this architecture can also be used for graph reordering.
Algorithm~\ref{alg:gnnpart-driver} summarizes the overall procedure.

\begin{algorithm}[!ht]
\caption{\our{}}
\label{alg:gnnpart-driver}
\begin{algorithmic}[1]
\Require Graph $G=(V,E)$, number of partitions $k$, GNN architecture $\mathcal{A}$, imbalance factors $\beta_{train}$, $\beta_{val}$,$\beta_{rest}$ migration policy $\mathcal{M}$
\Ensure Partition assignment $P: V \to \{1,\ldots,k\}$
\State $\mathbf{E} \gets \textsc{Phase1\_TrainAndEmbed}(G,\mathcal{A})$ \Comment{Alg.~\ref{alg:phase1}}
\State $P \gets \textsc{Phase2\_ClusterAndBalance}(\mathbf{E}, k, \beta_{train}, \beta_{val}, \beta_{rest}, \mathcal{M})$ \Comment{Alg.~\ref{alg:phase2}}
\State \Return $P$
\end{algorithmic}
\end{algorithm}

\paragraph{Phase 1: Embedding Computation.}
In Phase~1 (Algorithm~\ref{alg:phase1}: \textsc{TrainAndEmbed}), we execute the actual GNN training workload and extract a node embedding $e_v$ for each vertex $v$.
These embeddings encode both node features and structural information from the local neighborhood and serve as input to the subsequent clustering phase to partition the graph.

\begin{algorithm}[!ht]
\caption{Phase 1: \textsc{TrainAndEmbed}}
\label{alg:phase1}
\begin{algorithmic}[1]
\Require Graph $G=(V,E)$, GNN architecture $\mathcal{A}$
\Ensure Node embeddings $\mathbf{E}=\{e_v \mid v\in V\}$
\State Run the actual GNN workload on $G$
\State \Return $\mathbf{E}$ extracted from the GNN
\end{algorithmic}
\end{algorithm}

\paragraph{Phase 2: Clustering and Balancing.}

\begin{algorithm}[!ht]
\caption{Phase 2: \textsc{ClusterAndBalance}}
\label{alg:phase2}
\begin{algorithmic}[1]
\Require Embeddings $\mathbf{E}$, number of partitions $k$, imbalance factors $\beta_{train}$, $\beta_{val}$, $\beta_{rest}$, migration policy $\mathcal{M}$
\Ensure Partition assignment $P: V \to \{1,\ldots,k\}$
\State $P \gets \textsc{KMeans}(\mathbf{E}, k)$
\State $V_{train}^{(i)} \gets$ training vertices in partition $i$
\State $V_{val}^{(i)} \gets$ validation vertices in partition $i$
\State $V_{rest}^{(i)} \gets$ remaining vertices in partition $i$ 
\While{any |$V_{train}^{(i)}| > \beta_{train} \cdot \frac{|V_{\text{train}}|}{k}$}
  \State Migrate training vertices from overloaded to underloaded partitions using $\mathcal{M}(P, V_{train}, \beta_{train} \cdot \frac{|V_{\text{train}}|}{k})$
\EndWhile
\While{any |$V_{val}^{(i)}| > \beta_{val} \cdot \frac{|V_{\text{val}}|}{k}$}
  \State Migrate validation vertices from overloaded to underloaded partitions using $\mathcal{M}(P, V_{val}, \beta_{val} \cdot \frac{|V_{\text{val}}|}{k})$
\EndWhile
\While{any |$V_{rest}^{(i)}| > \beta_{rest} \cdot \frac{|V_{\text{rest}}|}{k}$}
  \State Migrate remaining vertices from overloaded to underloaded partitions using $\mathcal{M}(P, V_{rest}, \beta_{rest} \cdot \frac{|V_{\text{rest}}|}{k})$
\EndWhile
\State \Return $P$
\end{algorithmic}
\end{algorithm}

\begin{algorithm}[!ht]
\caption{Migration policy $\mathcal{M}$: Random migration proportional to free capacity}
\label{alg:migrate}
\begin{algorithmic}[1]
\Require Partition assignment $P$, candidate node set $\mathcal{S}$, capacities $C_i$ for each partition $i$
\Ensure Updated assignment $P$
\State Compute current loads $L_i = |\{v \in \mathcal{S} \mid P(v)=i\}|$
\State Identify overloaded partitions $O = \{i \mid L_i > C_i\}$ and underloaded partitions $U = \{i \mid L_i < C_i\}$
\State Compute free capacity $f_i = C_i - L_i$ for $i \in U$ and probabilities $p_i = f_i / \sum_{u \in U} f_u$
\For{each $i \in O$}
   \State Select $m_i = L_i - C_i$ nodes $i$ with smallest degree in $O$
   \For{each selected node $v$}
      \State Sample destination $i \in U$ with probability $p_i$
      \State Reassign: $P(v) \gets i$
   \EndFor
\EndFor
\State \Return $P$
\end{algorithmic}
\end{algorithm}

In Phase~2 (Algorithm~\ref{alg:phase2}: \textsc{ClusterAndBalance}), the node embeddings (one embedding per node) are clustered into $k$ groups using clustering.
We use \kmeans{} clustering; however, also other clustering approaches can be used.  
The resulting clusters are interpreted as graph partitions.  
While \kmeans{} always produces exactly $k$ partitions, it does not guarantee balance, and some partitions may contain significantly more vertices than others.  
Such imbalance can slow down distributed GNN training, as workers assigned to overloaded partitions become stragglers, forcing other workers to wait and overloaded partitions may exceed memory capacity on the assigned machine, leading to out-of-memory errors~\cite{gnnpartitioing}.

To address this, we apply a rebalancing procedure that migrates vertices from overloaded to underloaded partitions according to a migration policy $\mathcal{M}$ (Algorithm~\ref{alg:migrate}).  
We propose a highly scalable migration policy, though it can easily be replaced by more sophisticated alternatives. 
Our policy ensures that the vertex balance (see Equation~\ref{eq:vertex_balance}) does not exceed a given threshold $\beta$.  

First, we compute the maximum number of vertices allowed per partition such that the vertex-balance constraint is satisfied. 
Based on this limit, we identify the set of overloaded partitions $O$ and underloaded partitions $U$, and compute (i) the number of vertices that must be migrated from each partition in $O$ and (ii) the remaining capacity of each partition in $U$.  
For each partition in $O$, we then select the required number of vertices to migrate. Each selected vertex is assigned to a target partition in $U$ by sampling from $U$ with probability proportional to the available capacity.

One question is which vertices to migrate from overloaded partitions.  
In \our{}, the vertices with the smallest degree from each overloaded partition are selected.  
The intuition is that migrating low-degree vertices reduces the likelihood of increasing the edge cut, as these vertices are connected to fewer neighbors.
For example, migrating a vertex with degree three can only lead to three more cut edges, a high-degree vertex, however, can lead to many more cut edges.  

\subsection{Design Considerations and Extensions}
The architecture of \our{} offers several practical advantages that make it broadly applicable. 
We discuss these design considerations and extensions below.

\textbf{Modularity.}  
Each component of \our{}, embedding computation, clustering, and migration for balancing, can be independently replaced or enhanced. 
This modularity enables \our{} to be adaptable to various workloads and easily extended with \mbox{new techniques.} 

\textbf{Dynamic graphs.}  
\our{} naturally supports dynamic graphs.
Thanks to the inductive nature of GNN models, embeddings for newly added or modified nodes can be computed without requiring the retraining of the entire model. 
This property enables efficient partition updates in settings where the underlying graph evolves over time. 

\textbf{Re-partitioning.}  
\our{} also supports re-partitioning when system resources or training requirements change. 
Since partitioning is fast, the number of partitions can be adjusted dynamically, for example, to scale in or out. 
This flexibility is particularly important because memory consumption in GNN training depends strongly on the model architecture and hyperparameters (e.g., number of layers, feature size, and hidden dimension), and hyperparameter search may require different memory budgets across configurations, which in turn can change the number of machines needed.

\textbf{Model reuse and multi-model support.}  
Node embeddings generated with a specific GNN model can be used for partitioning and training other models. 
For example, embeddings obtained with GraphSAGE can be used to partition the graph for later training a graph attention network (GAT). 
Similarly, embeddings from a model trained for link prediction can be repurposed to partition graphs for node classification.

\textbf{Beyond partitioning: graph reordering.}  
Finally, \our{} can also be applied to \emph{graph reordering} for single-machine GNN training. 
Graph reordering improves cache locality and reduces training runtime, showing that \our{} has utility beyond distributed GNN training.

\section{Evaluation}
\label{sec:evaluation}
In the following, we evaluate \our{} with respect to partitioning quality metrics, partitioning run-time, and distributed GNN training performance. 
Then, we evaluate \our{} with regard to graph reordering quality metrics and single-machine GNN \mbox{training performance.} 

\textbf{Implementation.} 
We implemented \our{} with the state-of-the-art GNN system Deep Graph Library~(\dgl{})~\cite{dgl} to train GNN models (Phase 1) and use \kmeans{} clustering with Facebook AI Similarity Search (\textsc{FAISS})~\cite{douze2024faiss} to cluster node embeddings (Phase 2). 
\kmeans{} uses sampling-based clustering, meaning that the \textsc{K-Means} centroids are computed based on a sample of all embeddings. 
We found that $2^9$ embeddings per cluster lead to good quality.

\textbf{Baselines.}
We compare \our{} with state-of-the-art gaph partitioning approaches: We perform in-memory partitioning with \metis{}~\cite{metis} and \spinner{}~\cite{spinner}, streaming partitioning with \ldg{}~\cite{ldg}, buffered streaming partitioning with \cuttana{}~\cite{cuttana}, and random partitioning as a baseline.

\textbf{GNN architectures.}
We train two state-of-the-art models, GraphSage~\cite{sage} and Graph attention networks~(GAT)~\cite{velickovic2017graph}, as they are commonly used for distributed GNN training \cite{gnnpartitioing,compre:evaluation:vldb,P3}. 
We perform both node and link prediction tasks. 

\textbf{Evaluation platform.}
We use 4 servers equipped with an NVIDIA L40S GPU (Ada Lovelace Architecture, 48\,GB GDDR6 with ECC), an AMD EPYC GENOA 9454P 48-core processor, and 768\,GB of main memory each.

\textbf{Datasets.} We selected different real-world graphs for our evaluations: \AX{}, \PA{}, and \PR{} from the Open Graph Benchmark (OGB)~\cite{hu2020open} and \RE{} from the Deep Graph Library (DGL)~\cite{dgl}. The graphs are shown in Table~\ref{tab:evaluation:graphs} along with different graph properties. 

\begin{table}[!ht]
\caption{Evaluation graphs along with the number of vertices, number of edges, and mean degree.}
\label{tab:evaluation:graphs}
\begin{tabular}{|l|l|l|l|}
\hline
Graph &  Vertices &  Edges &  Mean Deg.  \\
\hline\hline
\AX{} &     169,343 &    1,166,243 &        13.77 \\
\RE{} &     232,965 &  114,615,892 &       983.98 \\
\PR{} &    2,449,029 &  123,718,280 &       101.03 \\
\PA{} &  111,059,956 & 1,615,685,872 &        29.10 \\
\hline
\end{tabular}
\end{table}

\textbf{Naming.}  
We emphasize that \our{} is an architecture for building embedding-driven graph partitioners, rather than a single partitioning algorithm. 
The actual partitioner instance depends on the underlying GNN architecture it was trained on. 
In the following, we use the naming convention: 
partitioners trained on \emph{link prediction} with two and three GNN layers are denoted as \otwol{} and \othreel{}, respectively, 
while partitioners trained on \emph{node (vertex) prediction} with two and three GNN layers are denoted as \otwov{} and \othreev{}, respectively. 

\subsection{When Are Embeddings Good Enough?}
A core advantage of \our{} is that it leverages the embeddings generated during the actual GNN training process, thus avoiding the cost of training a separate model solely for partitioning.
However, this raises an important question: \textit{After how many GNN training epochs is the model useful for producing high-quality partitions?}

To answer this, we investigate the partitioning quality (measured by edge-cut ratio and enforcing a vertex balance smaller 1.05) when using node embeddings generated after varying numbers of GNN training epochs, both for node and link prediction tasks. 
The results are shown in Table~\ref{tab:evaluation:epochs:cut-size:np} and \ref{tab:evaluation:epochs:cut-size:lp} for node and link \mbox{prediction, respectively.}
We summarize our key findings as follows:

\begin{table}[!ht]
\caption{Average edge-cut ratio (lower is better) after different numbers of GNN training epochs.}
\label{tab:evaluation:epochs:cut-size}
\centering
\begin{subtable}[t]{\columnwidth}
\centering
\begin{tabular}{|l|l|l|l|l|}
\hline
\#Epochs & \AX{} & \PA{} & \PR{} & \RE{} \\
\hline\hline
0 &        0.72 &             0.56 &           0.56 &    0.70 \\ \hline
1 &        0.70 &             0.49 &           0.51 &    0.52 \\ \hline
2 &        0.67 &             0.48 &           0.44 &    0.42 \\ \hline
3 &        0.61 &             0.49 &           0.39 &    0.37 \\ \hline
4 &        0.59 &             0.48 &           0.36 &    0.33 \\ \hline
5 &        0.56 &             0.49 &           0.34 &    0.31 \\ \hline
10 &        0.50 &             0.49 &           0.27 &    0.28 \\ \hline
30 &        0.45 &             0.49 &           0.22 &    0.27 \\ \hline
50 &        0.45 &             0.49 &           0.21 &    0.26 \\ \hline
70 &        0.44 &             0.49 &           0.20 &    0.27 \\ \hline
90 &        0.44 &             0.49 &           0.20 &    0.27 \\ \hline
\end{tabular}
\vspace{1em}
\caption{Node prediction.}
\label{tab:evaluation:epochs:cut-size:np}
\end{subtable}
\hfill
\begin{subtable}[t]{\columnwidth}
\centering
\begin{tabular}{|l|l|l|l|l|}
\hline
\#Epochs & \AX{} & \PA{} & \PR{} & \RE{} \\
\hline\hline
0 &                    0.61 &             0.51 &           0.55 &    0.70 \\\hline
1 &                    0.41 &             0.40 &           0.39 &    0.40 \\\hline
2 &                    0.38 &             0.39 &           0.27 &    0.33 \\\hline
3 &                    0.37 &             0.38 &           0.22 &    0.33 \\\hline
4 &                    0.36 &             0.37 &           0.19 &    0.31 \\\hline
5 &                    0.36 &             0.37 &           0.17 &    0.32 \\\hline
10 &                    0.35 &             0.35 &           0.15 &    0.29 \\\hline
30 &                    0.33 &             0.34 &           0.14 &    0.28 \\\hline
50 &                    0.32 &             0.34 &           0.14 &    0.28 \\\hline
70 &                    0.32 &             0.34 &           0.14 &    0.28 \\\hline
90 &                    0.32 &             0.35 &           0.15 &    0.29 \\\hline
\end{tabular}  
\vspace{1em}
\caption{Link prediction.}
    \label{tab:evaluation:epochs:cut-size:lp}
\end{subtable}
\end{table}
\begin{figure}[!ht]
  \centering
  \includegraphics[width=\linewidth]{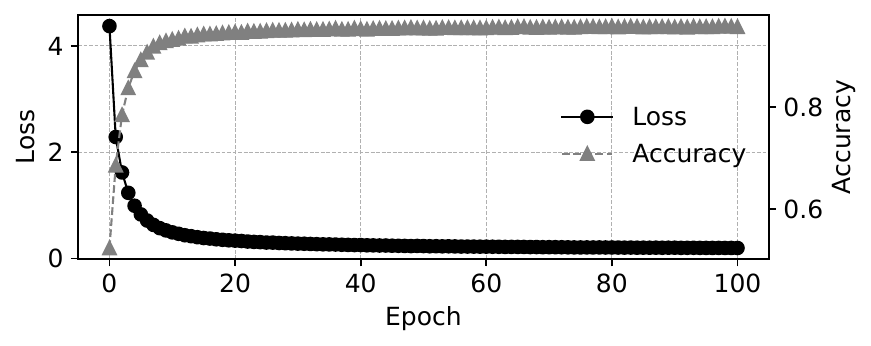}
  \caption{Loss and accuracy for \RE{} over epochs.}
  \label{fig:evaluation:epochs:loss:reddit}
\end{figure}

\textbf{(1) Partitioning quality improves with more training.}  
The edge-cut ratio decreases as the number of training epochs increases. 
After just 5 epochs, the quality is already quite good and after 30-50 epochs, it stabilizes and is close to the best achievable with our method. 
For example, the edge-cut ratio is heavily reduced for \PR{} from 0.56 to 0.21 for node prediction and from 0.55 to 0.14 for link prediction, respectively, when increasing the number of training epochs from 0 to 50. 

It is plausible that the edge-cut ratio is lower in the face of more training epochs. 
The more training epochs, the better the GNN in terms of prediction performance (accuracy), and also the more useful the embeddings produced by the model.
This is shown in Figure~\ref{fig:evaluation:epochs:loss:reddit} and was also observed for the remaining graphs: 
Even after a few training epochs, the validation loss decreases sharply and the accuracy of the GNN model is already quite high. 

However, at the beginning of our training workload, we do not have a trained GNN, meaning the parameters of the neural networks in the GNN are randomly initialized.  
In the following, we investigate how \our{} works with an untrained GNN. 

\textbf{(2) Even an untrained GNN (0 epochs) is beneficial.}  
Remarkably, using an untrained GNN to generate initial embeddings produces significantly better partitions than random partitioning. 
This suggests that GNN architectures encode inductive biases that help preserve graph structure, even before training.
Table~\ref{tab:evaluation:untrained:cut-size:np} and Table~\ref{tab:evaluation:untrained:cut-size:lp} show the reduction of the edge-cut ratio achieved by an untrained GNN in percent compared to random partitioning, for node prediction and link prediction, respectively. 
Across all datasets and partition counts, we observe significant improvements: compared to random partitioning, the edge-cut ratio decreases by between 7.03\% and 53.40\%.

\begin{table}[!ht]
\centering
\caption{Edge-cut ratio reduction by an untrained GNN compared to random partitioning in percentage. Larger is better.}
\label{tab:evaluation:untrained:cut-size}
\begin{subtable}[t]{0.48\textwidth}
\centering
\begin{tabular}{|l|l|l|l|l|}
\hline
\#Partitions & \AX{} & \PA{} & \PR{} & \RE{} \\
\hline\hline
2 &       10.55 &            49.91 &          37.90 &   19.19 \\\hline
4 &       14.81 &            39.07 &          37.49 &   22.55 \\\hline
8 &       13.53 &            30.42 &          32.63 &   10.69 \\\hline
16 &        9.73 &            23.14 &          25.82 &    8.71 \\\hline
32 &        7.03 &            19.25 &          22.30 &    9.69 \\\hline\hline
\textbf{Mean} & \textbf{11.13} & \textbf{32.36} & \textbf{31.23} & \textbf{14.17} 
\\\hline
\end{tabular}
\vspace{1em}
\caption{Node prediction.}
\label{tab:evaluation:untrained:cut-size:np}
\end{subtable}
\hfill
\begin{subtable}[t]{0.48\textwidth}
\centering
\begin{tabular}{|l|l|l|l|l|}
\hline
\#Partitions & \AX{} & \PA{} & \PR{} & \RE{} \\
\hline\hline
2 &       32.89 &            52.76 &          48.74 &   29.59 \\\hline
4 &       31.05 &            53.40 &          37.20 &   15.60 \\\hline
8 &       26.37 &            40.04 &          32.10 &   11.08 \\\hline
16 &       20.25 &            28.45 &          27.75 &    9.48 \\\hline
32 &       15.29 &            22.83 &          23.67 &    8.44 \\\hline\hline
\textbf{Mean} & \textbf{25.17} & \textbf{39.50} & \textbf{33.89} & \textbf{14.84} 
\\\hline
\end{tabular}
\vspace{1em}
\caption{Link prediction.}
\label{tab:evaluation:untrained:cut-size:lp}
\end{subtable}
\end{table}

These results point out a key strength of our method: even without additional training overhead, it can already provide substantial improvements over naive partitioning schemes. 
After just a few training epochs, the partitioning quality improves significantly, reaching competitive levels quickly.
Since the embeddings are obtained as a natural byproduct of the actual GNN workload, our method incurs virtually no extra cost, making it both efficient and practical for real-world use.

\subsection{Partitioning Performance}
\begin{figure}[!ht]
\centering
\begin{subfigure}{0.99\linewidth}
\centering
\includegraphics[width=\linewidth]{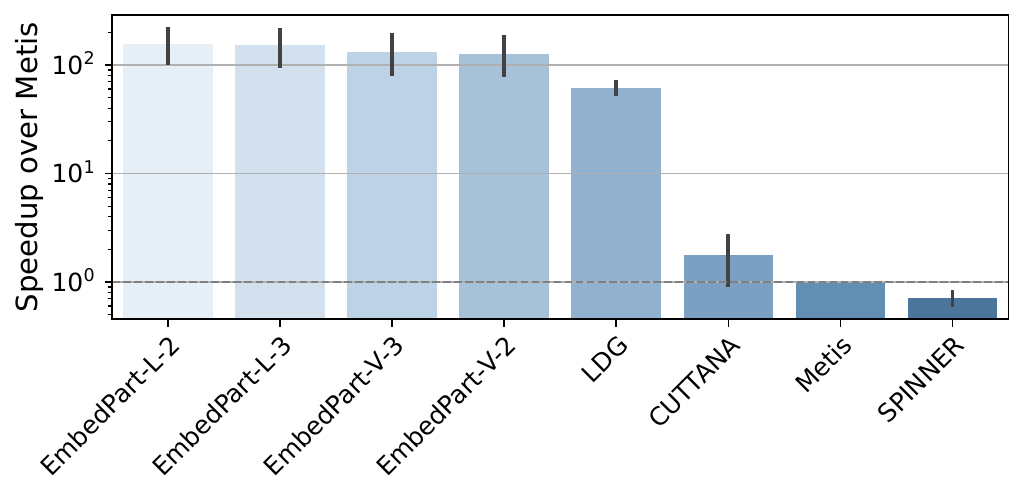}
\caption{Partitioning time speedup relative to \metis{}. Larger values indicate faster partitioning.}
\label{fig:partitioning:time:distribution}
\end{subfigure}
\begin{subfigure}{0.99\linewidth}
\centering
\includegraphics[width=\linewidth]{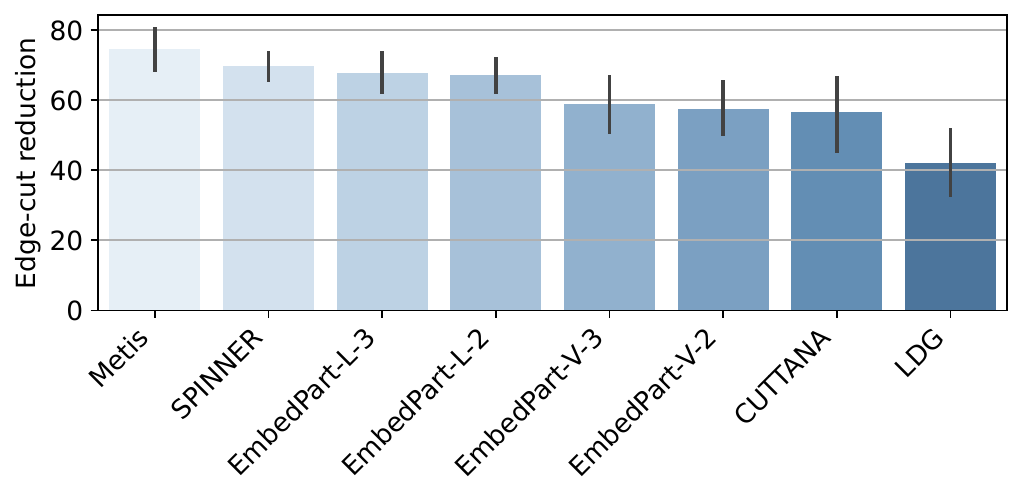}
\caption{Edge-cut reduction in \% relative to random partitioning. Larger values indicate better partitioning quality.}
\label{fig:partitioning:edge-cut:distribution}
\end{subfigure}

\caption{Comparison of partitioning time speedups and edge-cut reductions.}
\label{fig:partitioning:combined}
\end{figure}

We evaluate the partitioning performance of \our{} by partitioning the graphs listed in Table~\ref{tab:evaluation:graphs} into $k \in \{2, 4, 8, 16, 32\}$ partitions and measure partitioning time and different partitioning quality metrics. 
The main observations are as follows:

\textbf{(1) Significant reduction in partitioning time.}  
\our{} achieves drastic speedups over traditional in-memory partitioners and even outperforms streaming graph partitioning. 
Figure~\ref{fig:partitioning:time:distribution} gives an overview of the speedup of the partitioners over \metis{} over all graphs and the number of partitions. 

\begin{figure*}[!ht]
\centering
\begin{subfigure}[b]{\linewidth}
\centering
\includegraphics[width=\linewidth]{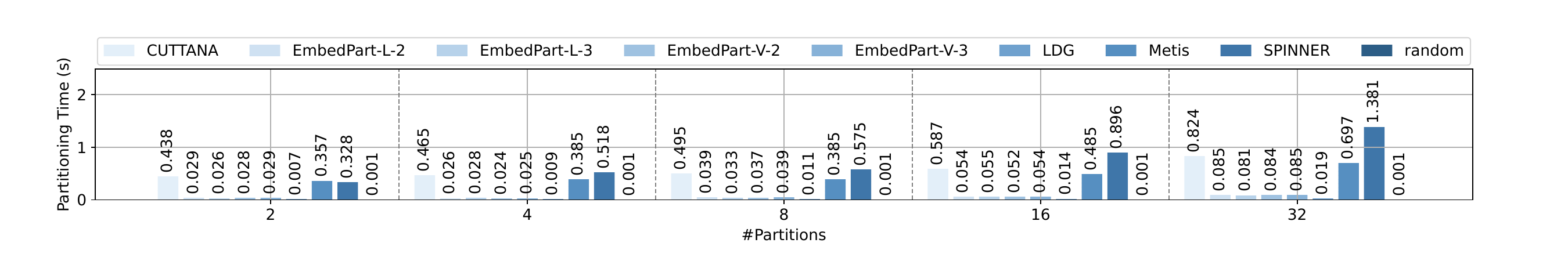}
\caption{\AX{}.}
\label{fig:partitioning-metrics:run-time:AX}
\end{subfigure}
\begin{subfigure}[b]{\linewidth}
\centering
\includegraphics[width=\linewidth]{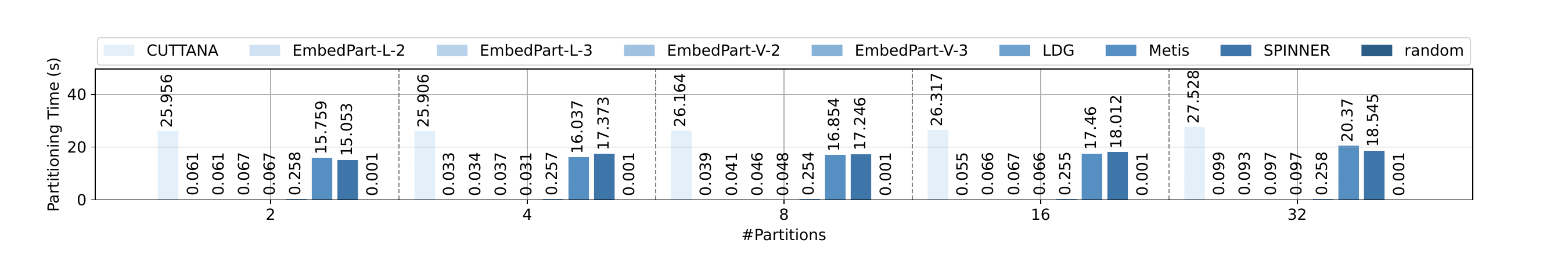}
\caption{\RE{}.}
\label{fig:partitioning-metrics:run-time:RE}
\end{subfigure} 
\centering
\begin{subfigure}[b]{\linewidth}
\centering
\includegraphics[width=\linewidth]{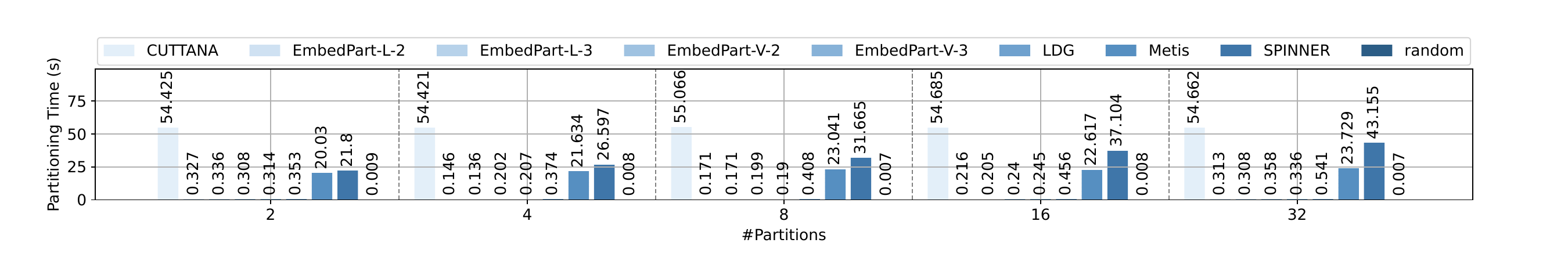}
\caption{\PR{}.}
\label{fig:partitioning-metrics:run-time:PR}
\end{subfigure} 
\centering
\begin{subfigure}[b]{\linewidth}
\centering
\includegraphics[width=\linewidth]{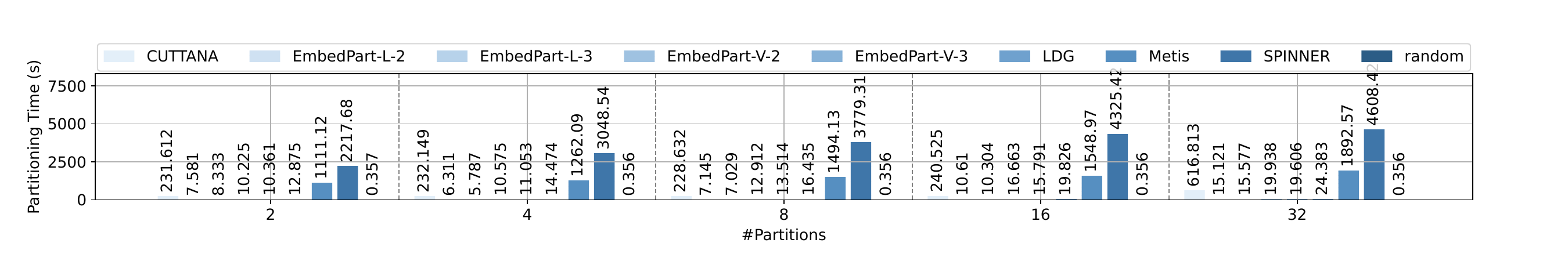}
\caption{\PA{}.}
\label{fig:partitioning-metrics:run-time:PA}
\end{subfigure} 
\centering
\caption{Partitioning run-time on different graphs and number of partitions. Lower is better}
\vspace{3mm}
\label{fig:partitioning-metrics:run-time}
\end{figure*}

Figures~\ref{fig:partitioning-metrics:run-time:AX}--\ref{fig:partitioning-metrics:run-time:PA} show all numbers per graph and the number of partitions.   
On average, \otwol{}, \othreel{}, \otwov{}, and \othreev{} achieve speedups between 127.12$\times$ and 155.27$\times$ over \metis{}. 
\spinner{} is significantly slower, achieving only a speedup of 0.71$\times$ (i.e., a slowdown) relative to \metis{}. 
Among the streaming approaches, \ldg{} and \cuttana{} achieve speedups of 61.10$\times$ and 1.78$\times$ over \metis{}, respectively.

Overall, \our{} exhibits the lowest partitioning time and is therefore well suited for dynamic or evolving graphs, where fast and repeated repartitioning is required.

\textbf{(2) Competitive partitioning quality.}  
Despite its speed, \our{} delivers strong partitioning quality.
It consistently achieves significantly lower edge cut ratios than random partitioning. 
Figure~\ref{fig:partitioning:edge-cut:distribution} shows how much the partitioners decrease the edge-cut ratio of random partitioning in percentages:
On average 
\metis{}, 
\spinner{}, 
\othreel{}, 
\otwol{}, 
\othreev{}, 
\otwov{}, 
\cuttana{},
and \ldg{}, 
reduce the edge-cut ratio by 
74.65\%,
69.81\%,
67.71\%,
67.14\%,
58.99\%,
57.49\%,
56.46\%, and
42.14\%
\othreel{} leads in 100\%, 80\%, 50\%, and 25\%, of all cases to a lower edge-cut ratio than \ldg{}, \cuttana{}, \spinner{}, and \metis{}, respectively. 

\begin{figure*}[!ht]
\centering
\begin{subfigure}[b]{0.99\linewidth}
\centering
\includegraphics[width=\linewidth]{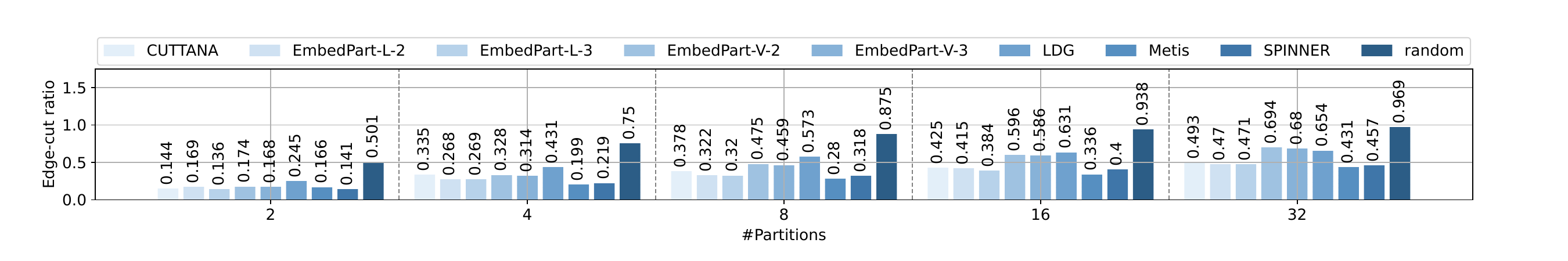}
\caption{\AX{}.}
\label{fig:partitioning-metrics:edge-cut:AX}
\end{subfigure}
\begin{subfigure}[b]{0.99\linewidth}
\centering
\includegraphics[width=\linewidth]{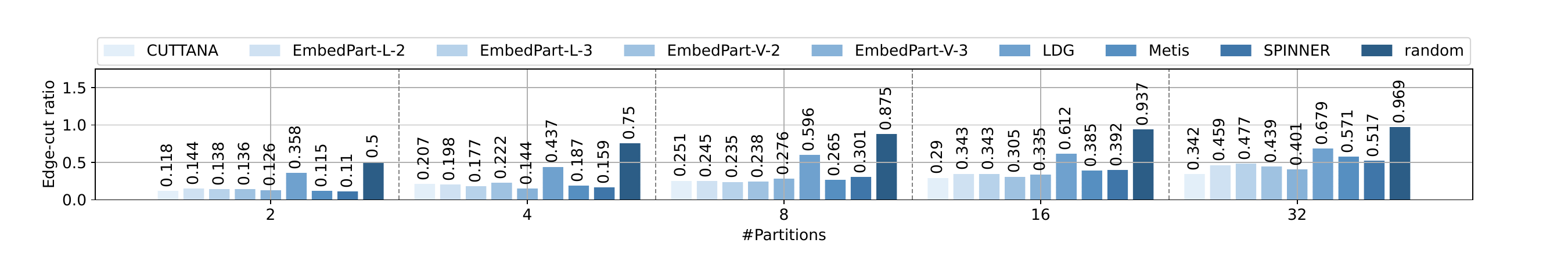}
\caption{\RE{}.}
\label{fig:partitioning-metrics:edge-cut:RE}
\end{subfigure} 
\centering
\begin{subfigure}[b]{0.99\linewidth}
\centering
\includegraphics[width=\linewidth]{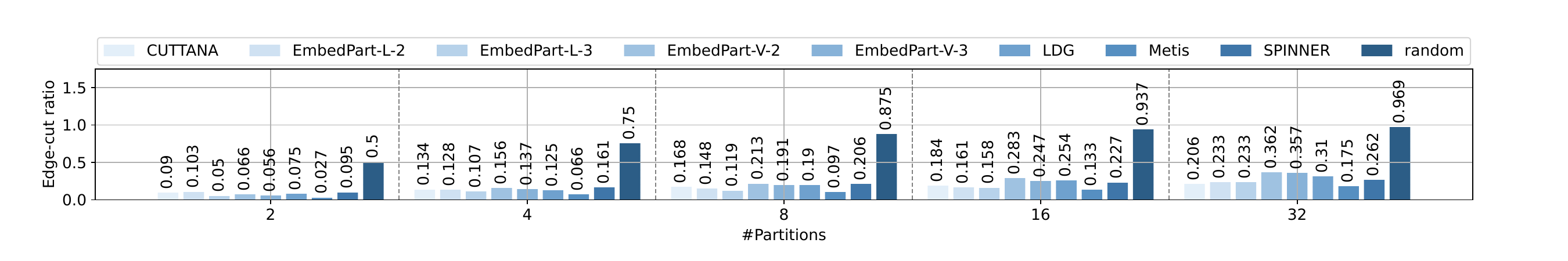}
\caption{\PR{}.}
\label{fig:partitioning-metrics:edge-cut:PR}
\end{subfigure} 
\centering
\begin{subfigure}[b]{0.99\linewidth}
\centering
\includegraphics[width=\linewidth]{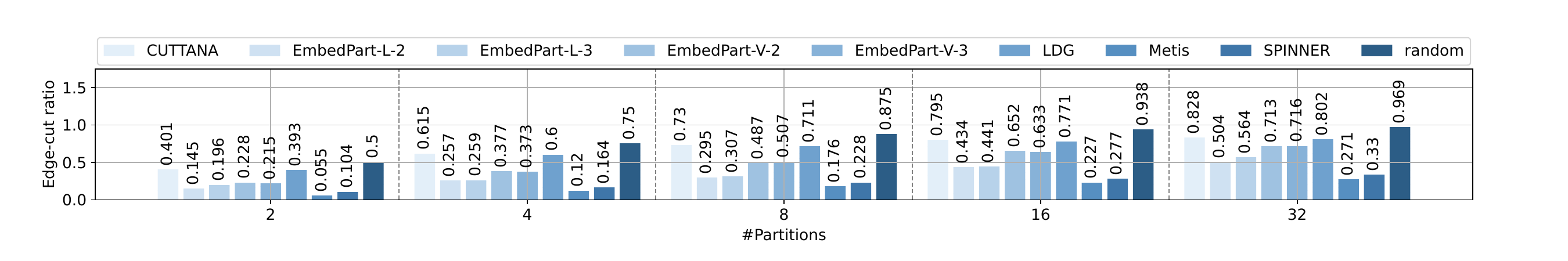}
\caption{\PA{}.}
\label{fig:partitioning-metrics:edge-cut:PA}
\end{subfigure} 
\centering
\caption{Partitioning quality metrics: Edge-cut ratio on different graphs and number of partitions. Lower is better.  }
\vspace{0.5em}
\label{fig:partitioning-metrics:edge-cut}
\end{figure*}

Therefore, \our{} is much faster in terms of partitioning time and competitive in terms of graph partitioning quality.  
For example, on the \RE{} dataset, \our{} outperforms both \metis{} and \spinner{} in certain configurations (see Figure~\ref{fig:partitioning-metrics:edge-cut:RE}).

Importantly, \our{} also provides an excellent balance of training vertices. 
It reliably respects the specified maximum imbalance ratio of $1.05$, while \cuttana{}, \ldg{} and \spinner{} often violate this constraint. 
For instance, on \AX{}, \cuttana{}, \ldg{} and \spinner{} yield training vertex imbalances of 1.4, 1.9 and 2.0 for 32 partitions, respectively and on \PR{}, they lead to imbalances of 1.9, 3.5 and 1.4 for 32 partitions, respectively. Similar trends are observed across the remaining datasets.

\textbf{We conclude that \our{} offers a good trade-off between partitioning speed and quality, making it a practical and scalable solution for partitioning large-scale graphs in modern distributed GNN training pipelines.}

\subsection{Distributed GNN Training Performance}
To evaluate the impact of graph partitioning on distributed GNN training performance, we run experiments on a cluster of four GPU servers. 
We use Distributed Deep Graph Library~(\texttt{DistDGL}) \cite{distdgl,dgl} to train 2- and 3-layer GAT and GraphSAGE models on all graphs listed in Table~\ref{tab:evaluation:graphs}, using 2 and 4 GPU servers. 
For graph partitioning, we use the same methods evaluated in the previous section: \metis{}, \spinner{}, \ldg{}, \cuttana{}, and our approach \our{}.

\begin{figure}[!ht]
\centering
\includegraphics[width=\linewidth]{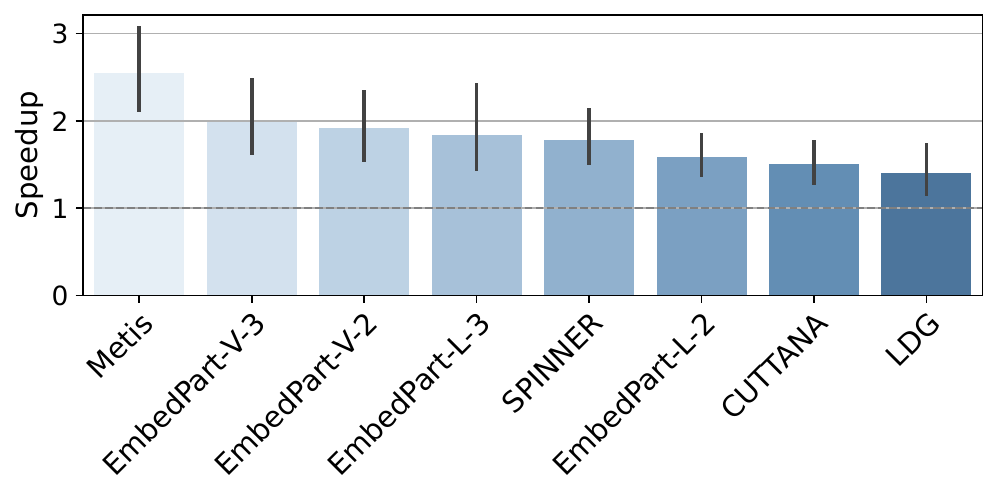}
\caption{Speedup of partitioners over random partitioning. Larger is better.}
\label{fig:speedup:allgraphs:distribution}
\end{figure}

\begin{figure}[!ht]
\centering
\begin{subfigure}[b]{0.95\linewidth}
\centering
\includegraphics[width=\linewidth]{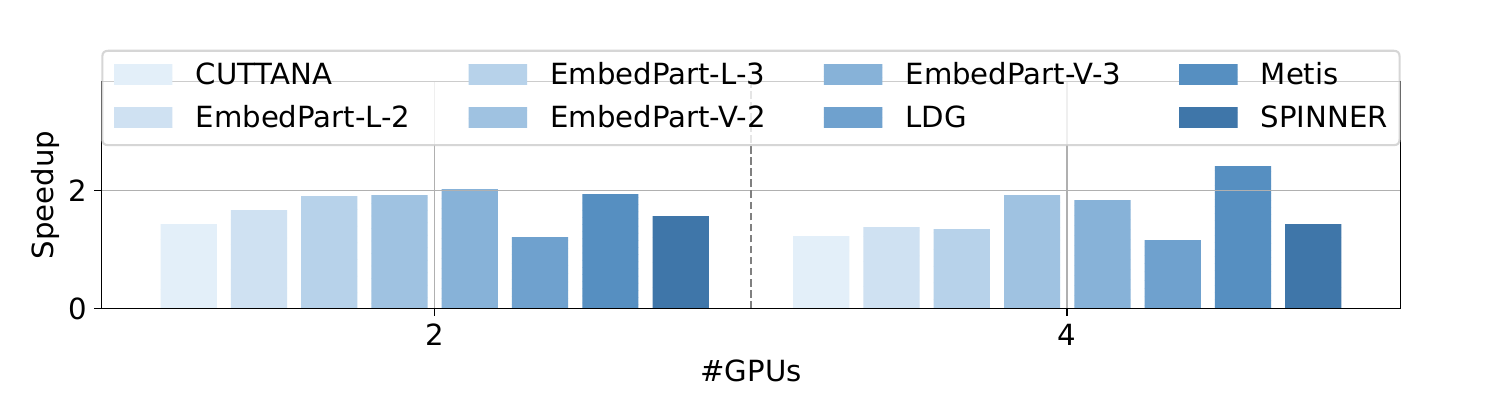}
\vspace{-7mm}
\caption{2-Layer GNN on \AX{}.}
\label{fig:dist:speedups:L2:AX}
\end{subfigure}
\begin{subfigure}[b]{0.95\linewidth}
\centering
\includegraphics[width=\linewidth]{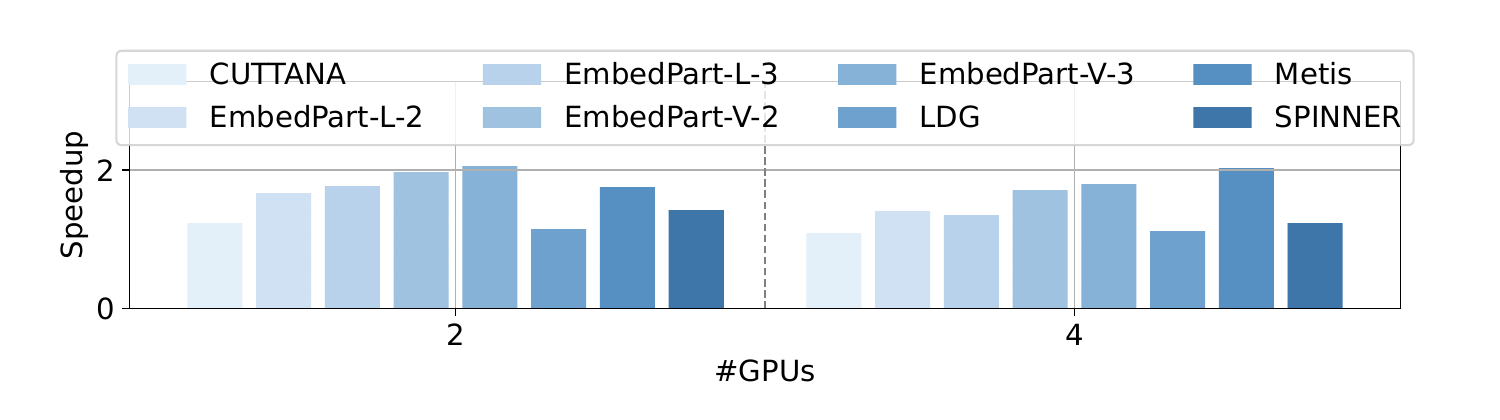}
\vspace{-7mm}
\caption{3-Layer GNN on \AX{}.}
\label{fig:dist:speedups:L3:AX}
\end{subfigure} 
\centering
\begin{subfigure}[b]{0.95\linewidth}
\centering
\includegraphics[width=\linewidth]{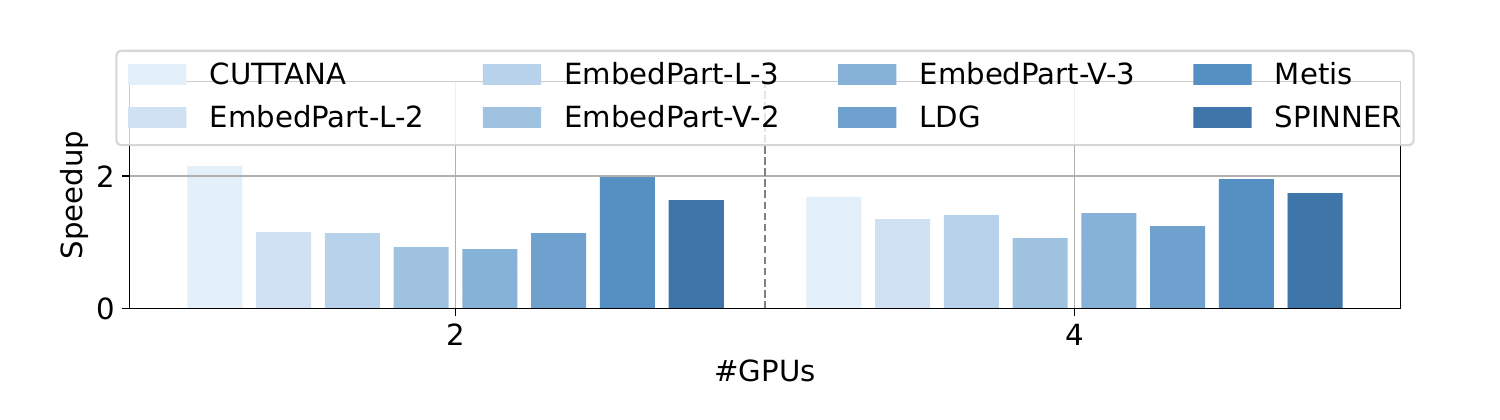}
\vspace{-7mm}
\caption{2-Layer GNN on \RE{}.}
\label{fig:dist:speedups:L2:RE}
\end{subfigure}
\begin{subfigure}[b]{0.95\linewidth}
\centering
\includegraphics[width=\linewidth]{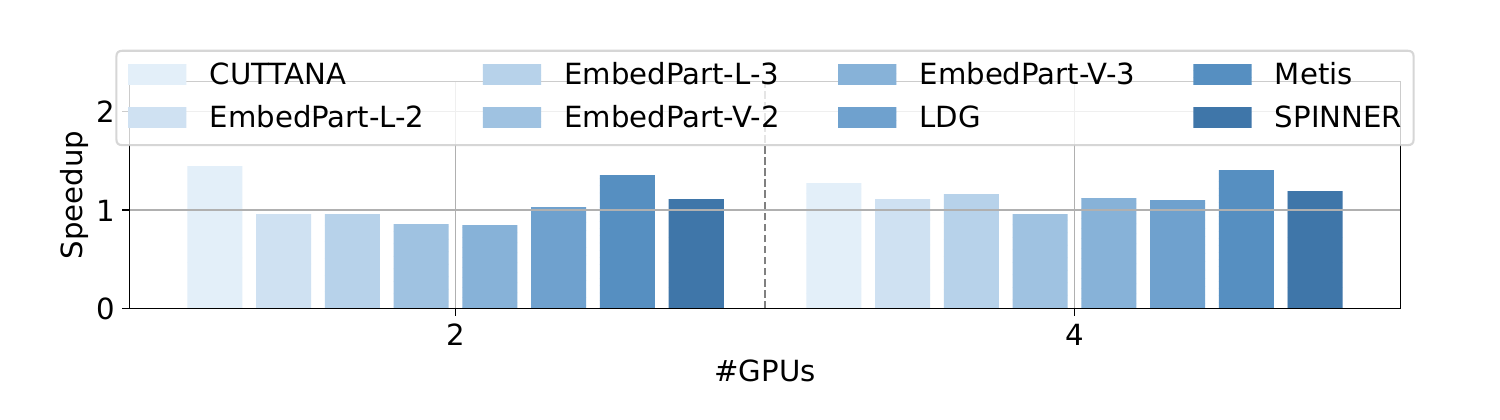}
\caption{3-Layer GNN on \RE{}.}
\vspace{-7mm}
\label{fig:dist:speedups:L3:RE}
\end{subfigure} 
\centering
\begin{subfigure}[b]{0.95\linewidth}
\centering
\includegraphics[width=\linewidth]{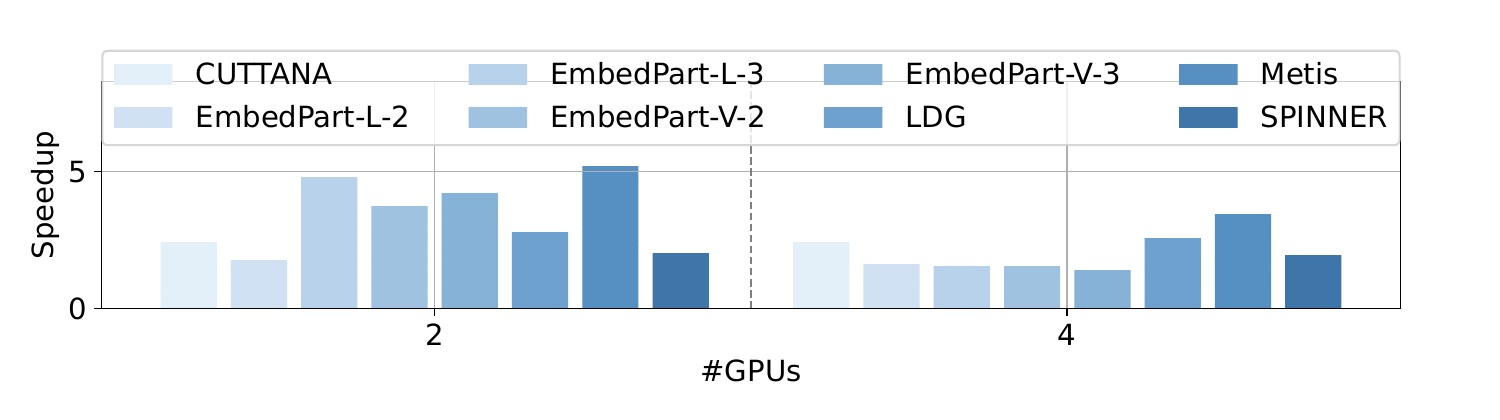}
\vspace{-7mm}
\caption{2-Layer GNN on \PR{}.}
\label{fig:dist:speedups:L2:PR}
\end{subfigure}
\begin{subfigure}[b]{0.95\linewidth}
\centering
\includegraphics[width=\linewidth]{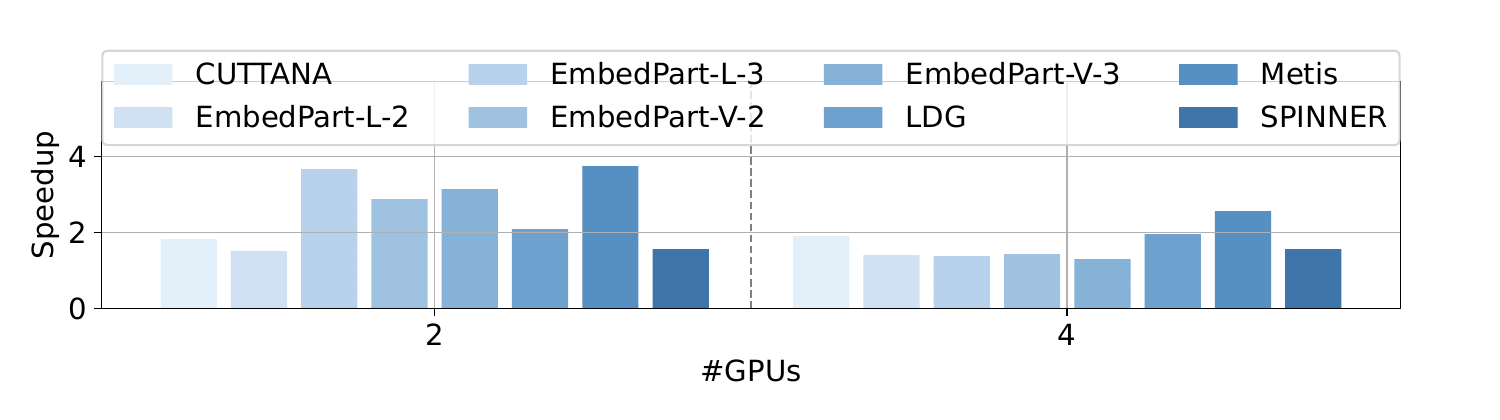}
\vspace{-7mm}
\caption{3-Layer GNN on \PR{}.}
\label{fig:dist:speedups:L3:PR}
\end{subfigure} 
\centering
\begin{subfigure}[b]{0.95\linewidth}
\centering
\includegraphics[width=\linewidth]{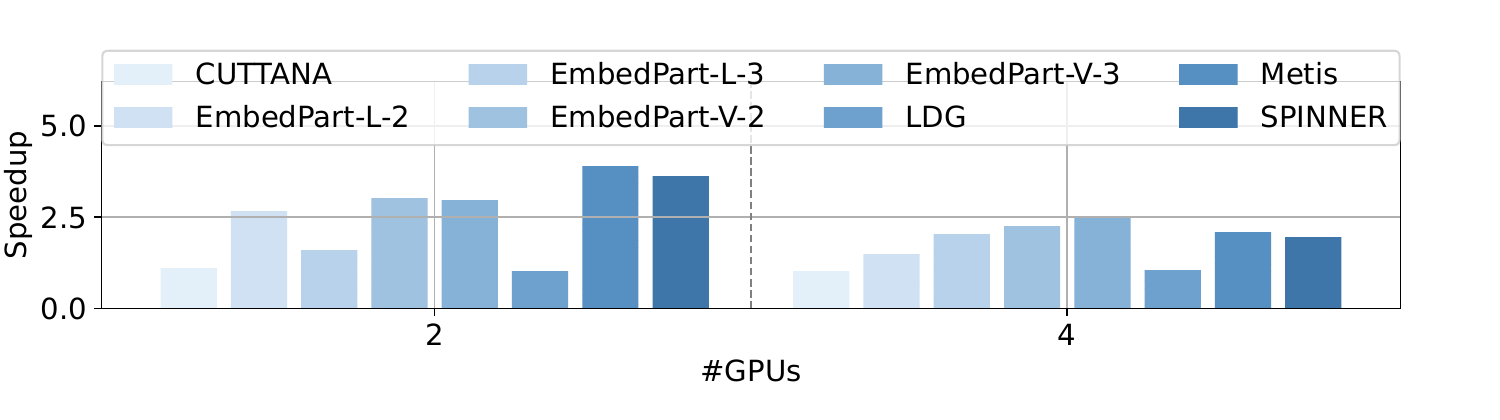}
\vspace{-7mm}
\caption{2-Layer GNN on \PA{}.}
\label{fig:dist:speedups:L2:PA}
\end{subfigure}
\begin{subfigure}[b]{0.95\linewidth}
\centering
\includegraphics[width=\linewidth]{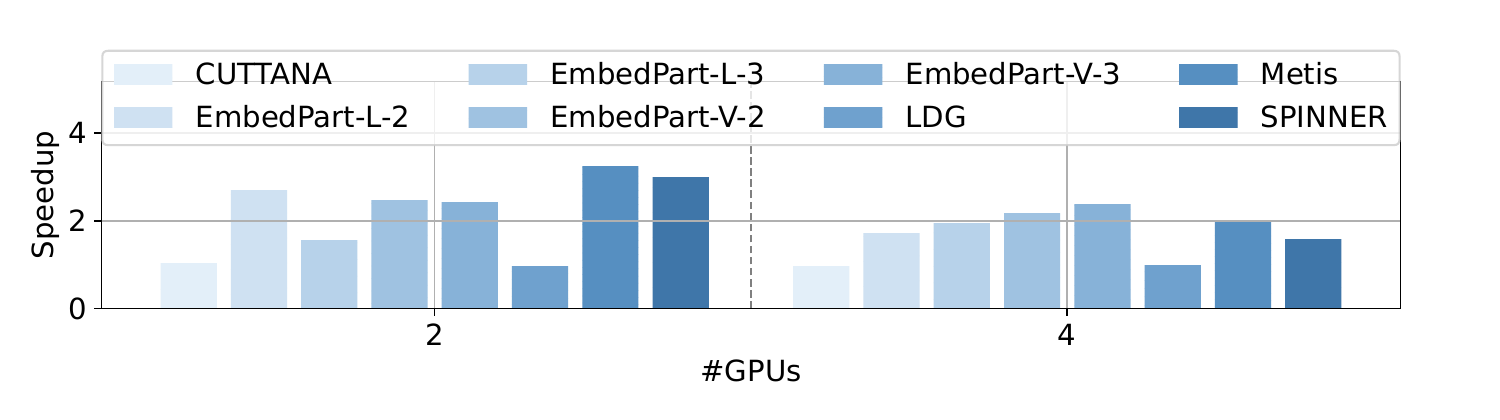}
\vspace{-7mm}
\caption{3-Layer GNN on \PA{}.}
\label{fig:dist:speedups:L3:PA}
\end{subfigure} 
\caption{Achieved speedups for distributed GNN training on different graphs, number of GPUs, and number of GNN layers. }
\label{fig:dist:speedups}

\end{figure}

We find that 
\metis{}, 
\othreev{}, 
\otwov{}, 
\spinner{}, 
\othreel{}, 
\otwol{},
\cuttana{}, and 
\ldg{} lead to average training speedups of 
2.55$\times$, 
2.01$\times$, 
1.92$\times$, 
1.84$\times$, 
1.78$\times$, 
1.59$\times$,
1.51$\times$, and 
1.41$\times$, 
respectively (see Figure~\ref{fig:speedup:allgraphs:distribution}). 
More detailed numbers are shown in Figures~\ref{fig:dist:speedups:L2:AX}-\ref{fig:dist:speedups:L3:PA} for each graph. 

We observe that \our{} leads to competitive distributed GNN training performance and outperforms \spinner{} and \ldg{} in many cases. There are even cases where \metis{} is outperformed.

These results underscore that \our{} provides an excellent trade-off between partitioning cost and training performance: it delivers \textbf{good training performance at a fraction of the partitioning time}, and substantially outperforms \ldg{} and \spinner{} in both speed and quality.

\subsection{Robustness to Graph Updates}
We now evaluate how well \our{} generalizes to updated graphs.
In many real-world scenarios, graphs evolve over time.
Let $G_t$ denote the state of the graph at time $t$. 
We have five subgraphs $G_1, \dots, G_5$ containing 10\%, 30\%, 50\%, 70\%, and 90\% of the vertices of the full graph $G$, respectively.

We train a GNN model for each graph $G_1, \dots, G_5$ and then apply each model directly to the full graph $G$ without any retraining to compute node embeddings for $G$, which are then used for graph partitioning. Then, we compare the achieved partitioning quality against a model that was trained directly on the full graph $G$.

Table~\ref{tab:evaluation:dynamicgraphs:cut-size} shows the resulting edge-cut ratios. 
Our key observation is that the \textbf{partitioning quality remains stable even as the graph changed significantly}. 
For example, on \PR{}, training on $G_2$ (only 30\% of $G$) yields an edge-cut of 0.23, which is very close to the 0.21 achieved when training directly on $G$. 
Similar trends hold across all datasets. 
While performance drops when training on $G_1$ (only 10\% of $G$), results become robust when training on 30\% or more, demonstrating the effectiveness of \our{} even under substantial graph updates.

This observation is plausible. GNNs are inductive models, meaning they are capable of generalizing to unseen vertices as long as the structural patterns and feature distributions are similar. 
As a result, when a model was trained on an older version of the graph it is often sufficient to achieve high-quality embeddings for partitioning the latest graph.

\begin{table}[!ht]
\caption{Average edge-cut ratio (lower is better) when training on different graph snapshots of $G$. The trained model is applied to the latest graph $G$ without retraining.}
\label{tab:evaluation:dynamicgraphs:cut-size}
\begin{tabular}{|l||l|l|l|l|}
\hline
Graph / size (\%) & \AX{} & \PA{} & \PR{} & \RE{} \\
\hline\hline
$G_1$ / 10&        0.56 &             0.50 &           0.30 &    0.32 \\\hline
$G_2$ / 30&        0.48 &             0.50 &           0.23 &    0.28 \\\hline
$G_3$ / 50&        0.46 &             0.49 &           0.21 &    0.27 \\\hline
$G_4$ / 70&        0.46 &             0.49 &           0.20 &    0.27 \\\hline
$G_5$ / 90&        0.45 &             0.49 &           0.20 &    0.26 \\\hline
$G$ / 100&        0.45 &             0.49 &           0.21 &    0.26 \\\hline
\end{tabular}
\end{table}

\subsection{Extension to Graph Reordering}
While \our{} is primarily designed for distributed GNN training, we also demonstrate how to use it as an effective optimization to accelerate \textit{single-machine GNN training} through \mbox{graph reordering.}

\begin{table*}[!ht]
\centering
\caption{Speedups through reordering for single-machine GNN training. Larger is better. OOM indicates out-of-memory errors.}
\label{tab:speedup:reordering}
\begin{tabular}{|l||llll|llll|}
\hline
 & \multicolumn{4}{c|}{CPU} & \multicolumn{4}{c|}{GPU} \\ \cline{2-9}
Strategy  &       \AX{} &    \PR{} &      \RE{}  &  \PA{} &      \AX{}  &   \PR{} &     \RE{} & \PA{} \\ \hline \hline
\cuttana{}   &           1.029 &           1.149 &           1.110 &           0.951 &           1.001 &           2.052 &  \textbf{1.026} &             OOM \\
\ldg{}       &           1.037 &           1.127 &           1.145 &  \textbf{1.022} &  \textbf{1.002} &           2.021 &           1.007 &             OOM \\
\metis{}     &           0.989 &           1.161 &           1.175 &           0.935 &  \textbf{1.002} &  \textbf{2.077} &           1.011 &             OOM \\
\spinner{}   &           1.003 &           1.127 &           1.136 &           0.955 &           1.000 &           2.018 &           1.011 &             OOM \\\hline
\othreel{}   &           1.025 &           1.147 &           1.147 &           0.954 &           1.001 &           1.995 &           1.013 &             OOM \\
\othreel{}-u &  \textbf{1.044} &  \textbf{1.165} &           1.186 &           1.017 &           1.000 &           2.074 &           1.022 &             OOM \\\hline
\othreev{}   &           1.002 &           1.109 &           1.147 &           1.016 &           0.999 &           1.948 &           1.024 &             OOM \\
\othreev{}-u &           0.993 &           1.148 &  \textbf{1.212} &           0.998 &           0.999 &           2.047 &           1.018 &             OOM \\\hline
\otwol{}     &           1.026 &           1.149 &           1.136 &           1.010 &           1.001 &           2.011 &           1.009 &             OOM \\
\otwol{}-u   &           1.011 &           1.162 &           1.171 &           0.995 &           1.001 &           2.073 &           1.021 &             OOM \\\hline
\otwov{}     &           1.024 &           1.134 &           1.146 &           0.990 &           1.001 &           1.994 &           1.009 &             OOM \\
\otwov{}-u   &           1.019 &           1.142 &           1.160 &           0.957 &           1.000 &           2.045 &           1.020 &             OOM \\\hline
\end{tabular}
\end{table*}

\begin{table*}[!ht]
\centering
\caption{Both the average graph bandwidth and edge-cut ratio are better (lower) for the unbalanced \our{}.}
\label{tab:reordering:quality}
\begin{tabular}{|l||llll|llll|}
\hline
 & \multicolumn{4}{c|}{Average graph bandwidth} & \multicolumn{4}{c|}{Edge-cut ratio} \\ \cline{2-9}
Strategy  &       \AX{} &    \PR{} &      \RE{}  &  \PA{} &      \AX{}  &   \PR{} &     \RE{} & \PA{} \\ \hline \hline
\othreel{}   &   74K &    1022K &   154K &     57594K  &      0.471 &         0.233 &  0.477 &           0.564 \\
\othreel{}-unbalanced &   56K &     659K &   124K &     38407K  &      0.360 &         0.162 &  0.372 &           0.286 \\\hline
\othreev{}   &   79K &    1196K &   151K &     61468K   &      0.680 &         0.357 &  0.401 &           0.716 \\
\othreev{}-unbalanced &   72K &     875K &   123K &     44247K  &      0.619 &         0.269 &  0.293 &           0.271 \\\hline
\otwol{}     &   71K &    1025K &   154K &     55970K  &      0.470 &         0.233 &  0.459 &           0.504 \\
\otwol{}-unbalanced   &   58K &     765K &   129K &     39413K  &      0.356 &         0.168 &  0.376 &           0.307 \\\hline
\otwov{}     &   79K &    1124K &   155K &     61169K  &      0.694 &         0.362 &  0.439 &           0.713 \\
\otwov{}-unbalanced   &   72K &     925K &   131K &     44830K  &      0.647 &         0.313 &  0.331 &           0.311 \\\hline
\end{tabular}
\vspace{0.5em}
\end{table*}

Graph reordering is a data layout optimization with the goal of improving data locality by placing vertices accessed together close by in memory. 
The IDs of vertices are used as the position in memory. 
In graph reordering the IDs are relabeled (reordered) in a way that vertices accessed together get similar IDs and therefore are stored close by in memory.
For GNN training, graph reordering is an effective because vertices iteratively aggregate information of neighboring vertices in the message passing phase (see Equation~\ref{eq:1}), meaning vertices access both features and intermediate embeddings of their neighbors~\cite{gnnreordering}.

According to \cite{gnnreordering}, the reordering quality metric that correlates the most with training speedup is the \textit{average graph bandwidth}, which we use for evaluation.
It is defined in the following. 
Let $\pi : V \rightarrow V$ be a bijection mapping vertices to their IDs. 
The \textit{gap} between two vertices $u, v \in V$ is defined as: 
\begin{equation}
\label{eq:gap}
\xi_{\pi}(u, v) = |\pi(u) - \pi(v)|
\end{equation}
The larger the gap between two vertices, the more distant they are from each other in memory. 
The \textit{vertex bandwidth} for a vertex $v \in V$ is defined as the distance to the most distant neighbor of $v$: 

\begin{equation}
\label{eq:vertex_bandwidth}
\beta_v(G,\pi)=\max \{ \xi_{\pi}(v, u) \mid \forall u \in n(v)\}
\end{equation}
Finally, the \textit{average graph bandwidth} of graph $G$ is defined as: 

\begin{equation}
\label{eq:agb}
\widehat{\beta}(G, \pi) = \frac{1}{|V|}\sum_{v \in V}\beta_v(G,\pi).
\end{equation}
The lower the average graph bandwidth the better the data locality in GNN training.

\subsubsection{Adapting \our{} for Reordering}
We adapt \our{} as follows to work for graph reordering: We reorder the graph such that vertices in the same partition get consecutive IDs.
The intuition is that vertices are densely connected within partitions, while connections between partitions are less likely.
This means that while a low edge-cut in distributed training reduces communication between machines, in single-machine training it reduces random memory accesses.

However, there is one crucial difference between single-machine and distributed GNN training scenarios. 
In distributed GNN training, vertex balancing ensures that every worker gets a similar share of the graph in terms of memory and workload, e.g., because features and intermediate representations must be stored in memory for each vertex. 
In contrast, this is not the case for single-machine GNN training. 
Therefore, we also evaluate \our{} without enforcing the vertex balance constraint, which was otherwise maintained at $\leq 1.05$ by our migration policy after the clustering phase.  
This has two advantages: First, we can skip the balancing step where vertices are migrated from overloaded to underloaded partitions. 
Second, the migration step reduces graph partitioning and graph reordering quality (edge-cut ratio and average graph bandwidth) because migrated vertices lead to higher edge-cut ratio and higher average graph bandwidth which will be shown in the following.

\subsubsection{Experimental Results}
In the following we analyse training speedups and graph reordering quality. 

\textbf{Training Speedup.}
Table~\ref{tab:speedup:reordering} reports the observed GNN training speedups achieved through graph reordering for CPU- and GPU-based training. 
We find that \our{} is effective for graph reordering in both CPU-based and GPU-based training.

For CPU-based training, \our{} achieves speedups up to 1.212$\times$, outperforming all baselines on \AX{}, \PR{}, \mbox{and \RE{}.}

For GPU-based training, we observe substantial improvements by graph reordering primarily on the \PR{} dataset.
Here, \metis{} achieves the highest speedup of 2.077$\times$, which is nearly identical to the 2.074$\times$ speedup achieved by \our{} (\otwol{}-unbalanced).
On the remaining datasets, GPU speedups are more modest, with most graph reordering strategies yielding only marginal~improvements.

\textbf{Reordering Quality.}
To understand the performance improvements, Table~\ref{tab:reordering:quality} shows both the average graph bandwidth and edge-cut ratio for balanced and unbalanced variants of \our{}. 
We find that removing the balance constraint drastically improves reordering quality (lower is better for both~metrics).

For average graph bandwidth on \PR{}, the unbalanced versions reduce from 1022K, 1196K, 1025K, and 1124K to 659K, 875K, 765K, and 925K for \othreel{}, \othreev{}, \otwol{}, and \otwov{}, respectively, representing improvements of 35\%, 27\%, 25\%, and 18\%.
A similar trend is observed across all~graphs.

Similarly, we observe substantial decreases in edge-cut ratio on \PR{} from 0.233, 0.357, 0.233, and 0.362 to 0.162, 0.269, 0.168, and 0.313 for \othreel{}, \othreev{}, \otwol{}, and \otwov{}, respectively.

Furthermore, the unbalanced variants can significantly reduce partitioning time.
For instance, the partitioning time of \othreel{} and \othreel{}-unbalanced is 15.6s and 7.8s on \PA{}, respectively, corresponding to a 50\% reduction in preprocessing overhead while simultaneously improving reordering quality since the migration phase is skipped and only clustering is performed.

\textbf{Takeaway.}  
Although \our{} is not optimized specifically for graph reordering, it achieves the best performance in most cases for CPU-based GNN training, while being substantially faster than competing methods.
This demonstrates the broader applicability and versatility of our embedding-based partitioning approach beyond distributed GNN training alone. 
It also highlights the advantages of our modular architecture, where we can replace or skip different phases depending on the use case. For example, completely skipping the migration phase for balancing is effective for single-machine GNN training where balance is not as critical as in distributed GNN training, resulting in both better quality partitions and reduced preprocessing time.

\section{Discussion}

\textbf{Embedding Reuse in Iterative ML Workflows.}
Model development in machine learning typically involves systematic hyperparameter optimization and model selection, which require evaluating multiple configurations of architectures~\cite{bergstra2012random,snoek2012practical}.
In the context of GNNs, exploring different layer depths, hidden dimensions, or training parameters can substantially affect memory consumption and computational demand, potentially changing the number of machines required for distributed training~\cite{gnnpartitioing}.
In addition, models are often retrained due to data drift~\cite{10.1145/2523813}.
Consequently, GNN training is rarely a one-shot process.
\our{} is designed for such iterative settings.
Once available, node embeddings can be reused to derive updated partitionings, enabling fast repartitioning when scaling resources up or down or retraining on updated data.

\textbf{Flexible Balancing Policies.}
The migration-based balancing step is lightweight and modular.
Balancing objectives can be adapted by assigning weights to vertices, and different imbalance thresholds can be incorporated without modifying the overall architecture.
In our evaluation, we follow common practice in graph partitioning literature and enforce balance with respect to the number of vertices~\cite{10.1145/3571808}. 

\section{Related Work}
\label{sec:relatedwork}
\textbf{Graph Partitioning for Distributed Processing.}
Graph partitioning is a long-standing and vibrant area of research for optimizing distributed graph processing.
A wide range of partitioning algorithms has been proposed \cite{ldg,spinner,cusp,hep,metis,cuttana,twops,gcnsplit}.

Many works have shown that graph partitioning can speed up distributed graph processing workloads such as PageRank, Shortest Paths, Connected Components~\cite{survey.1.vldb.2017,survey.2.vldb.2018,survey.3.vldb.2018,survey.4.sigmod.2019,cut.to.fit,ease}.
More recently, it has also been demonstrated that graph partitioning can substantially accelerate distributed GNN training~\cite{gnnpartitioing}. 

Our work fundamentally differs from prior partitioning work.
Existing methods rely on graph structure, which is highly irregular and difficult to process efficiently and treat partitioning as a separate pre-processing step.
We instead partition based on dense node embeddings that are naturally generated as part of the actual GNN training workload which can be performed highly efficiently.
We view this embedding-driven perspective as a new research direction that enables further opportunities for optimization.

\textbf{Graph Reordering for Single-machine Processing.}
Graph reordering is a data management optimization that improves data locality by placing frequently accessed vertices close together in memory, thus improving cache utilization.
Prior work has shown its effectiveness for single-machine graph processing of classical analytics workloads \cite{when,gorder}, and more recently for single-machine GNN training \cite{gnnreordering,dgi}. 

In our work, we demonstrate that \our{} can also be applied to graph reordering, further broadening its applicability.

\textbf{System for Graph Neural Network Training.}
Many specialized systems~\cite{AliGraph,distdgl,dgl,sancus,NeutronTP,DistGNN,P3,G3,NeutronTask,ginex,Seastar,NeutronStar,diskgnn,sheng2024outre,hongtu,mariusgnn} have been proposed to support GNN training at scale. For our evaluation, we build upon one of the most widely used open-source systems, which also serves as the foundation for several subsequent GNN systems \cite{DistGNN,P3}.

\textbf{Data Management Techniques for GNN Training.}
A complementary line of work investigates data management optimizations for GNN workloads~\cite{datamgmt4GNN,compre:evaluation:vldb,ijcai2022p772}, including sampling to reduce computation and communication overheads by training on subgraphs~\cite{sage,ADGNN,samplingsurvey,clustergcn,insitu,ringsampler}, caching of frequently accessed data to mitigate I/O bottlenecks~\cite{pagraphcaching,AliGraph,gids,ginex}, and compression and sparsification to reduce memory overhead and accelerate message passing~\cite{featurecompression,adaptivesparsify,demystifysparsification,graphcompression}.

In contrast, we propose \our{}, an embedding-driven architecture for graph partitioning and reordering, providing a fresh perspective on these core data management optimizations.

\section{Conclusions}
\label{sec:conclusions}

We presented \our{}, an embedding-driven architecture for graph partitioning that shifts partitioning from graph topology to the dense embedding space produced during GNN training.
By leveraging node embeddings that are already generated as part of the training workload, \our{} transforms graph partitioning into a scalable clustering problem over dense representations.

Our evaluation shows that this design reduces partitioning time by more than two orders of magnitude compared to state-of-the-art in-memory partitioners while maintaining competitive partitioning quality and distributed GNN training performance.
At the same time, the approach naturally supports dynamic graphs, fast repartitioning, and can also be applied to graph reordering to accelerate single-machine GNN training.

More broadly, our work highlights the potential of embedding-driven data management techniques that leverage learned representations to guide system-level optimizations.
As machine learning workloads increasingly operate on learned embeddings, these representations provide new opportunities for designing scalable and adaptive data layout strategies.

\begin{acks}
This work is funded in part by the Deutsche Forschungsgemeinschaft (DFG, German Research Foundation) - 438107855.
\end{acks}

\bibliographystyle{ACM-Reference-Format}
\bibliography{sample-base}

\end{document}